\documentclass[aps, physrev, twocolumn, groupedaddress]{revtex4-2}
\usepackage[braket]{qcircuit}
\usepackage{bm}
\usepackage{dcolumn}
\usepackage{float}
\usepackage{graphicx}
\usepackage{subcaption}
\usepackage{amsmath}
\usepackage{xcolor}
\usepackage{placeins}
\usepackage{hyperref}
\usepackage{cleveref}
\usepackage{tabularray}
\usepackage{csquotes}
\UseTblrLibrary{booktabs}
\UseTblrLibrary{siunitx}
\Crefname{figure}{Fig.}{Figs.}
\Crefname{equation}{Eq.}{Eqs.}
\makeatletter
\AddToHook{cmd/appendix/before}{\def\cref@section@alias{appendix}}
\makeatother

\graphicspath{{./}}
\begin{document}
    \title{A Versatile Variational Quantum Kernel Framework for Non-Trivial Classification}
    \author{Jiang Yuhan}
    \affiliation{Department of Physics, University of Wisconsin--Madison, Madison, WI 53706, USA}

    \author{Matthew Otten}
    \email{mjotten@wisc.edu}
    \affiliation{Department of Chemistry, University of Wisconsin--Madison, Madison, WI 53706, USA}
    \affiliation{Department of Physics, University of Wisconsin--Madison, Madison, WI 53706, USA}

    \date{October 13, 2026}

    \begin{abstract}
        Quantum kernel methods are a promising branch of quantum machine learning, but their performance across data types and encoding choices remains incompletely understood. To address this gap, we developed an algorithmic framework for variational quantum kernels utilizing resource-efficient ansätze for complex classification tasks and introduced a parameter scaling technique to accelerate convergence. We conducted a comprehensive benchmark of this framework on eight challenging, real-world and high-dimensional datasets covering tabular, image, time series, and graph data. Our results show that the proposed quantum kernel demonstrated competitive classification accuracy compared to standard classical kernels in classical simulation, such as the radial basis function (RBF) kernel. This work demonstrates that properly designed quantum kernels can function as versatile, high-performance tools, laying a foundation for quantum-enhanced applications in machine learning. Further research is needed to fully assess the practical performance of quantum methods.
    \end{abstract}

    \maketitle

    \section{Introduction}
    Quantum kernel methods have shown promise and are gaining growing use among quantum machine learning approaches to enhance the performance of kernel-based models, where support vector machines (SVMs) are a common example \cite{wangComprehensiveReviewQuantum2024}. They have been applied to various machine learning tasks, such as classification of medical data or high-energy physics \cite{flotherHowQuantumComputing2025, incudiniAutomaticEffectiveDiscovery2024}. An advanced enhancement to these kernel methods is the trainable quantum kernel, which employs a parameterized quantum circuit (PQC), often referred to as an ansatz. Here, a quantum circuit's gate operations are controlled by a set of externally optimized classical parameters  \cite{peruzzoVariationalEigenvalueSolver2014, farhiQuantumApproximateOptimization2014}. This enables the quantum kernel to be trained and adapted to the specific structure of a dataset \cite{schuldQuantumMachineLearning2019}.

    However, despite theoretical promise, the practical deployment of quantum kernel methods is still in its very early stages. Many research studies focus on a single specific machine learning area with a few dataset samples, but an evaluation of the performance of a quantum kernel across diverse domains remains unverified, whereas this ability is common in classical kernel methods such as the linear kernel or Radial Basis Function (RBF) kernel \cite{fernandez-delgadoWeNeedHundreds2014}. This makes it difficult to understand the characteristics of the methods' performance from a comprehensive perspective. Furthermore, existing practice is primarily conducted on low-dimensional synthetic or introductory datasets like variants of MNIST or Iris, or aggressively reduced real-world data that goes from hundreds or more to around ten features \cite{schnabelQuantumKernelMethods2025,alvarez-estevezBenchmarkingQuantumMachine2025, miroszewskiSearchQuantumAdvantage2024}. Large-scale benchmarking studies further caution that small QML evaluations are highly design-sensitive and that out-of-the-box classical models often match or exceed fixed, untrained quantum kernels \cite{bowlesBetterClassicalSubtle2024}. Recent work remains largely single-domain or limited in training protocol: medical imaging with features pooled to 49--64 dimensions \cite{singhBenchmarkingMedMNISTDataset2026}, financial prediction where an untrained fidelity kernel fell short of classical RBF \cite{ahmadQuantumVsClassical2026}, tabular QSVM benchmarks with aggressive dimension reduction where kernel--target alignment (KTA) training is competitive only in isolated cases \cite{kakavandBenchmarkingQuantumKernel2026}, and neural quantum kernels that pretrain a quantum neural network rather than aligning a variational embedding by KTA \cite{rodriguez-grasaNeuralQuantumKernels2025}. The regime of KTA-trained quantum kernels under cross-domain evaluation with minimal dimension reduction therefore remains open. Designing and training these circuits on classical simulators also introduces practical challenges, such as convergence issues, that are not well addressed by current methods.

\begin{figure}[htbp]
        \includegraphics[width=\columnwidth]{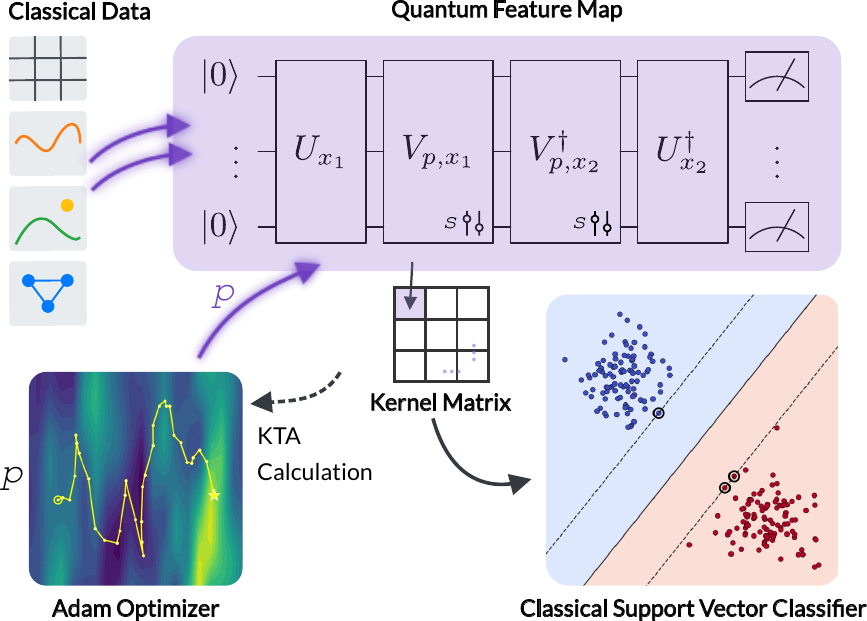}
        \caption{Overview of the variational quantum kernel framework. Classical data is encoded into quantum states via a parameterized quantum circuit (Quantum Feature Map), which incorporates a scaling parameter $s$ on the variational gates. The overlaps of these states form a Kernel Matrix. The circuit's parameters $p$ are trained using a classical optimizer (e.g., Adam \cite{kingmaAdamMethodStochastic2017}) to maximize the Kernel-Target Alignment (KTA). The final, optimized kernel is used by a Classical Support Vector Classifier for classification.\label{fig:overview}}
    \end{figure}

    The aforementioned limitations---the lack of cross-domain validation, reliance on oversimplified datasets, and the absence of structured circuit design and practical QML techniques for classical simulation---hinder our ability to assess the real-world viability of quantum kernel methods. To address these gaps, this work develops a comprehensive and robust quantum kernel methodology, designed and validated for high-dimensional, complex classification problems as shown in \Cref{fig:overview}, where a trainable quantum circuit encodes data into a kernel matrix that is optimized through Kernel-Target Alignment and used by a Support Vector Classifier for classification \cite{hubregtsenTrainingQuantumEmbedding2022}.

    We use both quantum-native and classical-inspired circuit architectures that are resource-efficient ansätze built on $N$ qubits and repeated for $L$ layers. They require only $\log(d)$ qubits for $d$-dimensional data or even as few as 2, and each layer is composed of single-qubit rotations and a circular chain of CNOT gates for entanglement, resulting in a total CNOT count that scales as $\mathcal{O}(LN)$ and a circuit depth of $\mathcal{O}(LN)$. To accelerate convergence, we introduce a parameter scaling technique that controls the expressivity of the ansatz, enabling a high-quality initial embedding and effective training. Then, we conduct a comprehensive evaluation of quantum kernels, rigorously testing eight datasets across tabular, image, time series, and graph types in medicine, physics, chemistry, biology and computer science areas with minimal and logical sample and dimension reduction. This way, we demonstrate that properly designed quantum kernels can exceed classical kernel performance across diverse data types while operating within small resource requirements and establishing an important foundation for their practical application in machine learning pipelines. Our results, obtained through classical simulation, not only validate the current potential of quantum kernel methods but also uncover new avenues for future research as the hardware constraints are gradually lifted. Future directions include extending our framework to more complex, multimodal data and exploring more quantum-native approaches, such as structure-aware feature maps, as an alternative to representing data as numerical vectors.

    \section{Kernel Design \label{kernel_design}}

    The design of a quantum kernel, which defines a similarity measure between data points in a quantum feature space, involves several critical choices within its underlying PQC. Recent systematic studies on hybrid quantum-classical models provide critical guidance on this matter. Specifically, a comprehensive study on hybrid convolutional neural networks provides a practical framework for understanding the relative importance of a PQC's core components: data encoding, the variational ansatz, and measurement \cite{lozano-cruzPracticalInsightsEffect2025}.

    The study revealed a clear hierarchy of influence, identifying the data encoding as the most significant factor, with its choice leading to performance variations exceeding 30\%. In contrast, the specific structure of the variational ansatz and the measurement basis were found to have a comparatively marginal effect. This hierarchy is directly relevant to quantum kernel methods, which are themselves a form of hybrid quantum-classical algorithm. We can establish the following correspondence:

    \begin{itemize}
        \item Data Encoding: Corresponds to the quantum feature map, implemented by a unitary encoder $\mathcal{U}(\textbf{x})$ that maps a classical data point $\textbf{x}$ to a quantum state $|\psi(\textbf{x})\rangle = \mathcal{U}(\textbf{x})|0\rangle^{\otimes n}$. As established, the design of this feature map is the dominant factor determining the kernel's expressive power and subsequent classification performance.
        \item Variational Ansätze: Corresponds to the optional, trainable secondary unitary $\mathcal{V}(\theta)$ that can be appended to perform techniques like quantum kernel alignment \cite{glickCovariantQuantumKernels2024}. While this ansatz introduces learnable parameters, its structural choice has been shown to have a less critical impact than the initial data encoding.
        \item Measurement: The kernel entry is typically calculated using the fidelity kernel, defined by the squared inner product $K_{ij} = |\langle\psi(\textbf{x}_i)|\psi(\textbf{x}_j)\rangle|^2$. Unlike the general PQC case where the measurement basis is a design hyperparameter, the measurement objective for a quantum kernel is algorithmically fixed. This value is typically estimated using a standard method, such as the invert-and-measure technique shown in \Cref{fig:generic_circuit}.
    \end{itemize}

    Given this prior evidence, our work systematically explores the design of the data encoding, as both theory and recent empirical results suggest this is the most impactful lever for optimizing a quantum kernel's performance.

    \subsection{Data encoding}
    \subsubsection{Amplitude Encoding}
    Currently, the primary quantum encoding methods are basis encoding, angle encoding, and amplitude encoding. It has been demonstrated that there was no difference in the predictive performance among these three quantum encoding methods \cite{rathQuantumDataEncoding2024}. The first two methods require $d$ qubits to encode data that has $d$ features, while amplitude encoding requires only $\lceil\log_2 d\rceil$. For the purpose of encoding data which can easily have hundreds to millions of features \cite{golubMolecularClassificationCancer1999,krizhevskyImageNetClassificationDeep2017,chengWideDeepLearning2016}, the former approach is significantly inefficient and would make the application of such machine learning tasks very difficult and impractical. Therefore, we use the amplitude encoding as one of our encoding methods that is quantum-native.

    To implement this encoding, we require a vector of length $2^n$; therefore, we first pad our input vectors to the target dimension by cyclically repeating their elements. This method expands the vector to the nearest power of two by repeatedly appending the original feature set, which, unlike zero-padding, ensures that amplitude is not allocated to uninformative zero-states. The resulting padded vector is then L2-normalized, and its components are embedded as the amplitudes of the computational basis states,
    \begin{equation} |\psi_x\rangle = \sum_{i=0}^{2^n-1} x'_i |i\rangle. \end{equation}

    \subsubsection{Truncated RBF Encoding}
    Among common classical kernel methods, the RBF kernel is widely considered the state-of-the-art choice for general-purpose applications due to its ability to handle non-linear relationships and its strong empirical performance across a wide variety of datasets \cite{hsuPracticalGuideSupport2003}. We use this as the foundation of the data encoding, employing a truncated version of the RBF encoding inspired by previous studies \cite{ottenQuantumMachineLearning2020,liuQuantumKerrLearning2023}. The method leverages the properties of canonical coherent states, $|\alpha\rangle$, whose squared overlap yields the form of the classical RBF kernel.

    We encode a classical data feature $x$ into the coherent state parameter $\alpha$ using a trainable length-scale hyperparameter $c$, such that $\alpha = x / (\sqrt{2}c)$. This is directly equivalent to the RBF kernel length-scale $l$ and relates to the $\gamma$ parameter via $c = \sqrt{\frac{1}{2\gamma}}$. To implement this on a system with a finite $D$-dimensional Hilbert space (corresponding to $\log_2(D)$ qubits), the feature state $|\phi(x)\rangle$ is prepared using the first $D$ terms of the canonical coherent state's Fock basis expansion,
    \begin{equation} |\phi(x)\rangle = C \sum_{n=0}^{D-1} \frac{\alpha^n}{\sqrt{n!}} |n\rangle, \end{equation}
    where $C$ is a normalization constant. This approach creates a feature map
    based on a mathematical truncation of the ideal coherent state vector. While
    this differs from the operator-based truncation and the Trotterization method, it serves as an efficient, simulation-based approximation of a coherent state kernel.

    Based on preliminary analysis, here we use two qubits that form a four-dimensional Hilbert space, which we found to be sufficient for the approximation kernel to approach the performance of the classical RBF kernel on our chosen datasets.

    We will denote the kernel using Amplitude Encoding as \texttt{QAmp} and the kernel using Truncated RBF Encoding as \texttt{QRBF} in the following sections.

    \subsection{Variational Ansätze}
    Following the encoding, an ansatz built on $N$ qubits and repeated for $L$ layers is applied.

    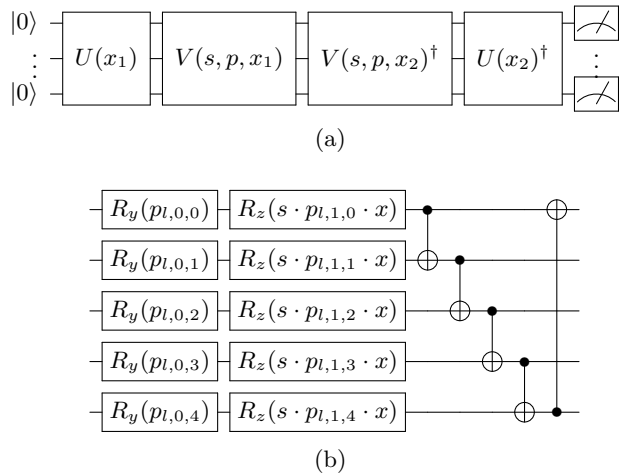
\begin{figure}[htbp]
        \begin{subfigure}{\columnwidth}
        \makebox[\columnwidth][c]{
            \Qcircuit @C=0.5em @R=0.3em {
            \lstick{\ket{0}} & \multigate{2}{U(x_1)} & \multigate{2}{V(s, p, x_1)} & \multigate{2}{V(s, p, x_2)^\dagger} & \multigate{2}{U(x_2)^\dagger} & \meter \\
            \lstick{\vdots} & \ghost{U(x_1)} & \ghost{V(s, p, x_1)} & \ghost{V(s, p, x_2)^\dagger} & \ghost{U(x_2)^\dagger} & \vdots \\
            \lstick{\ket{0}} & \ghost{U(x_1)} & \ghost{V(s, p, x_1)} & \ghost{V(s, p, x_2)^\dagger} & \ghost{U(x_2)^\dagger} & \meter
            }
        }
        \caption{\label{fig:generic_circuit}}
        \end{subfigure}

        \hfill

        \begin{subfigure}{\columnwidth}
        \makebox[\columnwidth][c]{
        \Qcircuit @C=0.5em @R=0.5em {
            \lstick{} & \gate{R_y(p_{l,0,0})} & \gate{R_z(s \cdot p_{l,1,0} \cdot x)} & \ctrl{1} & \qw      & \qw      & \qw      & \targ     & \qw \\
            \lstick{} & \gate{R_y(p_{l,0,1})} & \gate{R_z(s \cdot p_{l,1,1} \cdot x)} & \targ    & \ctrl{1} & \qw      & \qw      & \qw       & \qw \\
            \lstick{} & \gate{R_y(p_{l,0,2})} & \gate{R_z(s \cdot p_{l,1,2} \cdot x)} & \qw      & \targ    & \ctrl{1} & \qw      & \qw       & \qw \\
            \lstick{} & \gate{R_y(p_{l,0,3})} & \gate{R_z(s \cdot p_{l,1,3} \cdot x)} & \qw      & \qw      & \targ    & \ctrl{1} & \qw       & \qw \\
            \lstick{} & \gate{R_y(p_{l,0,4})} & \gate{R_z(s \cdot p_{l,1,4} \cdot x)} & \qw      & \qw      & \qw      & \targ    & \ctrl{-4} & \qw \\
            }
        }
        \caption{\label{fig:ansatz}}
        \end{subfigure}
        \caption{(a) Circuit diagram for estimating the kernel entry $K(x_1, x_2) = |\langle \psi(x_2) | \psi(x_1) \rangle|^2$. The feature map $|\psi(x) \rangle$ is generated by the data-encoding unitary $U(x)$ followed by the trainable variational block $V(s, p, x)$. (b) Structure of a single ansatz layer built on five qubits. Each wire undergoes a parameterized $R_y$ rotation and an $R_z$ rotation scaled by the data point and a hyperparameter $s$, followed by a circular chain of CNOT gates for entanglement.}
    \end{figure}

    Each layer $l$ consists of two parts, as shown in \Cref{fig:ansatz}. First, a layer of single-qubit gates is applied to all qubits, where each qubit $i$ undergoes an $R_Y(\theta_{i,l})$ rotation, followed by an $R_Z(\phi_{i,l} \cdot x)$ rotation. The parameters $\theta_{i,l}$ and $\phi_{i,l}$ are trainable, and the re-uploading of the input $x$ in each layer makes the overall ansatz $V(x)$ data-dependent, a necessary condition for creating a non-trivial ansatz. For \texttt{QAmp}, it reuploads the scalar mean of the feature vector $x$, while for \texttt{QRBF}, it reuploads the $x$ scalar directly. Second, an entangling block of CNOT gates is applied in a circular chain, with CNOTs acting between adjacent qubits from $i$ (control) to $(i+1) \pmod N$ (target). This entire layer structure is repeated $L$ times. The circuit has a total of $2 \cdot N \cdot L$ trainable parameters.

    To ensure training stability and control the model's expressivity, we introduce a hyperparameter, $s$, that globally scales the argument of each data-dependent $R_Z$ gate. The rotation is thus applied as $R_Z(s \cdot \phi_{i,l} \cdot x)$. This scaling factor is not optimized during training but is tuned through hyperparameter search for each dataset. This technique improves training stability and accelerates convergence.

    In terms of quantum resources, the complete $L$-layer ansatz requires a total of $LN$ CNOT gates and $2LN$ single-qubit gates. The depth of the ansatz scales linearly with both the number of layers and qubits, with a total depth of $L(N+2)$, excluding the initial state preparation. The complete circuit involves an ansatz and its adjoint, and has a total depth of $2L(N+2)$. For the $L=5$ layers used in all our experiments, this corresponds to $5N$ CNOT gates, $10N$ single-qubit gates, and a final circuit depth of $10(N+2)$.

    \section{Datasets}
    To comprehensively assess the performance of our quantum kernels, we create a benchmark of eight datasets. This selection follows the hierarchy of data complexities that have historically defined the development of classical kernel methods, spanning four canonical categories of machine learning challenges and five scientific domains, as illustrated below.

    \subsubsection{Tabular Data}
    Tabular data represents the foundational use case for which Support Vector Machines were originally conceived \cite{cortesSupportvectorNetworks1995}. As these datasets are inherently structured as feature vectors, they provide a clean baseline to test the kernel's core effectiveness in a standard Euclidean feature space.

    \begin{itemize}
        \item TCGA-LGG: Sourced from medicine, this dataset contains gene expression data (20530 features) for 511 Lower-Grade Glioma patients \cite{cancergenomeatlasresearchnetworkComprehensiveIntegrativeGenomic2015}. The binary classification task is to distinguish between two primary molecular subtypes based on their genomic profile.
        \item Higgs Boson: From the domain of high-energy physics, this dataset consists of 250000 CERN proton-proton collision events described by 30 kinematic features \cite{baldiSearchingExoticParticles2014,adam-bourdariosHiggsBosonMachine2014}. The task is the binary classification of events into either a Higgs boson signal or background noise.
        \item QSAR Biodegradation: This chemical informatics dataset contains 1055 molecules described by 41 quantitative structure-activity relationship (QSAR) descriptors \cite{mansouriQuantitativeStructureactivityRelationship2013}. The binary classification task is to predict whether a compound is readily biodegradable or not.
    \end{itemize}

    \subsubsection{Image Data}
    Image data represents the canonical challenge of high-dimensionality. Before the dominance of convolutional networks, a leading approach involved flattening images into high-dimensional vectors and applying kernel SVMs \cite{csurkaVisualCategorizationBags2002}.

    \begin{itemize}
        \item Fashion-MNIST: A widely-used computer vision benchmark \cite{xiaoFashionMNISTNovelImage2017}, Fashion-MNIST consists of 70000 grayscale images (28x28 pixels) of ten apparel classes. For our experiments, we focus on the binary classification task of distinguishing between classes zero (T-shirt/top) and six (Shirt), which is the hardest pair for the classical kernels according to preliminary analysis. We will refer to this data as the Fashion-Tshirt/Shirt.
    \end{itemize}

    \subsubsection{Time Series Data}
    Time series data introduces the challenge of sequential dependency. This domain challenges a kernel's ability to model temporal patterns, a weakness in standard kernels that prompted the development of specialized temporal kernels \cite{cuturiKernelTimeSeries2007}.

    \begin{itemize}
        \item SEED-EEG: From the domain of affective computing, we use features extracted from multi-channel (62-channel) EEG time series signals from the SEED dataset \cite{zhengInvestigatingCriticalFrequency2015,duanDifferentialEntropyFeature2013}. The task is the three-class classification of the subject's emotional state (positive, neutral, or negative). Our experiments focus on a single session (\texttt{12\_20131127}, the first session of the participant 12), which contains 15 trials (film clips) processed into 3394 feature vectors. We refer to this experimental subset as SEED-P12S1.
        \item PhysioNet Challenge 2017: This cardiology dataset contains 8528 single-lead Electrocardiogram (ECG) time series of varying lengths \cite{cliffordAFClassificationShort2017}. The objective is the four-class classification of heart rhythms into normal sinus rhythm, atrial fibrillation, an alternative rhythm, or a noisy recording. For our study, we focused on the binary task of distinguishing between normal sinus rhythm and atrial fibrillation. This data will be referred to as PhysioNet2017-NA.
    \end{itemize}

    \subsubsection{Graph Data}
    Finally, graph-structured data pushes kernel methods to their modern frontier. Standard kernels fail on these non-Euclidean graphs, a problem that gave rise to the entire subfield of graph kernels, with the Weisfeiler-Lehman (WL) kernel being a prominent state-of-the-art example \cite{shervashidzeWeisfeilerLehmanGraphKernels2011}.

    \begin{itemize}
        \item MUTAG: A standard benchmark in chemoinformatics accessed from the TUDataset collection \cite{morrisTUDatasetCollectionBenchmark2020}, this dataset consists of 188 graph-structured chemical compounds \cite{debnathStructureactivityRelationshipMutagenic1991}. The binary classification task is to predict the mutagenic effect of each compound based on its atomic structure.
        \item PROTEINS: A bioinformatics dataset available through the TUDataset collection, comprising 1113 graphs where nodes represent secondary structure elements of proteins \cite{borgwardtProteinFunctionPrediction2005}. The binary classification task is to determine whether each protein is an enzyme or a non-enzyme.
    \end{itemize}

    \section{Experimental Pipeline\label{sec:pipeline}}
    \subsubsection{Data Preprocessing}
    To compare quantum and classical kernel models, we established a preprocessing procedure using the scikit-learn library \cite{pedregosaScikitlearnMachineLearning2011}. For reproducibility, all internal operations involving stochastic processes, such as data splitting, were executed with the run seed.

    The dataset was first loaded and cleaned according to the classical standard ML process, depending on the condition of the data. For datasets other than the tabular type, feature engineering into vectors is necessary for the kernel method to process the data, if the original source data has not already been processed accordingly; for graphs this is aggregation of node attributes and basic size and degree statistics. Then, the data is split randomly into a training set of 75\% and a test set of 25\%, with stratification for imbalanced datasets using \texttt{train\_test\_split}. Fashion-MNIST was already split to a train set of 60000 samples and a test set of 10000 samples, and SEED-P12S1 is split with nine trials as training set and six trials as test set, as described in the dataset paper.

\begin{figure}[htbp]
        \begin{subfigure}{\columnwidth}
            \includegraphics[width=\linewidth]{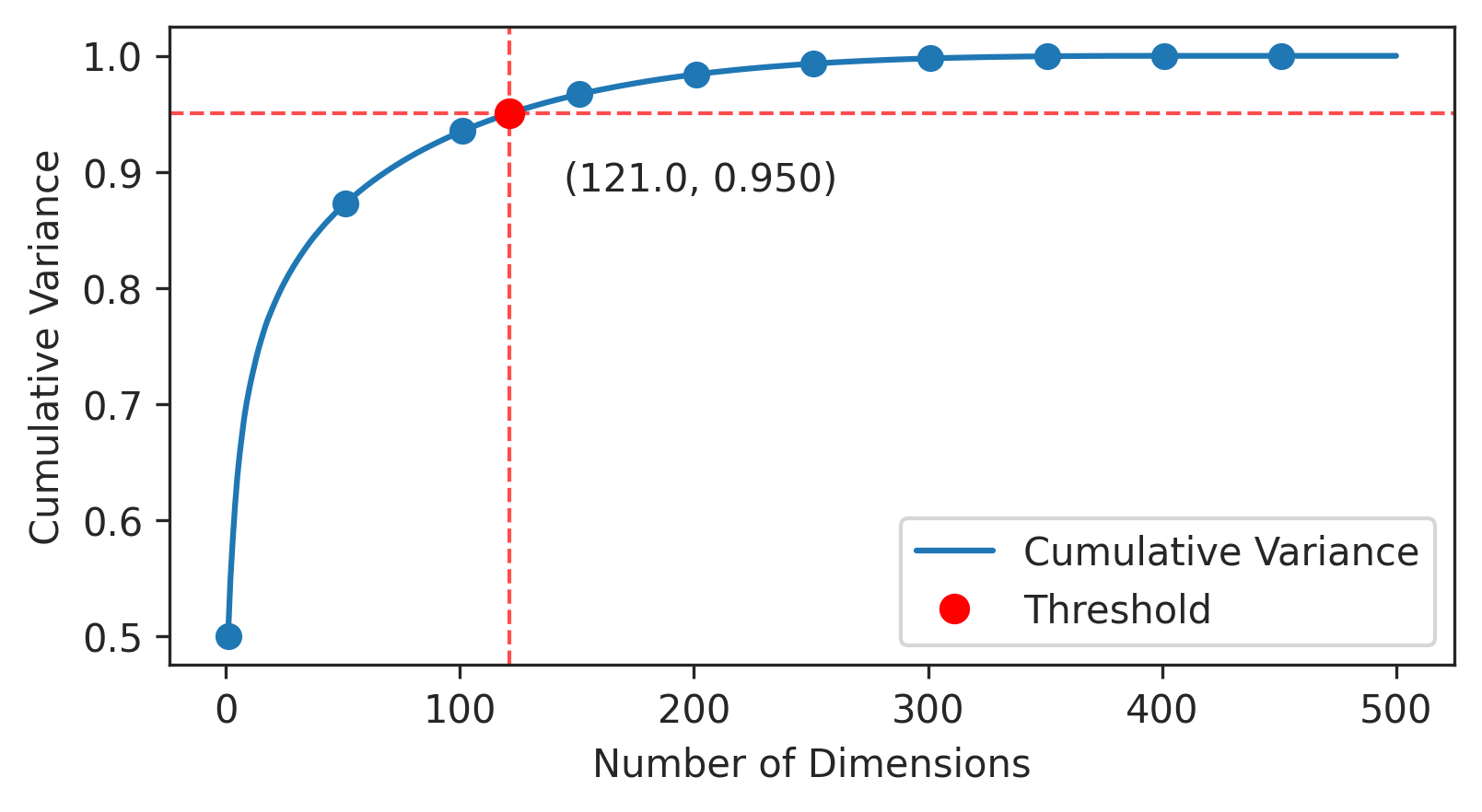}
        \end{subfigure}

        \begin{subfigure}{\columnwidth}
            \includegraphics[width=\linewidth]{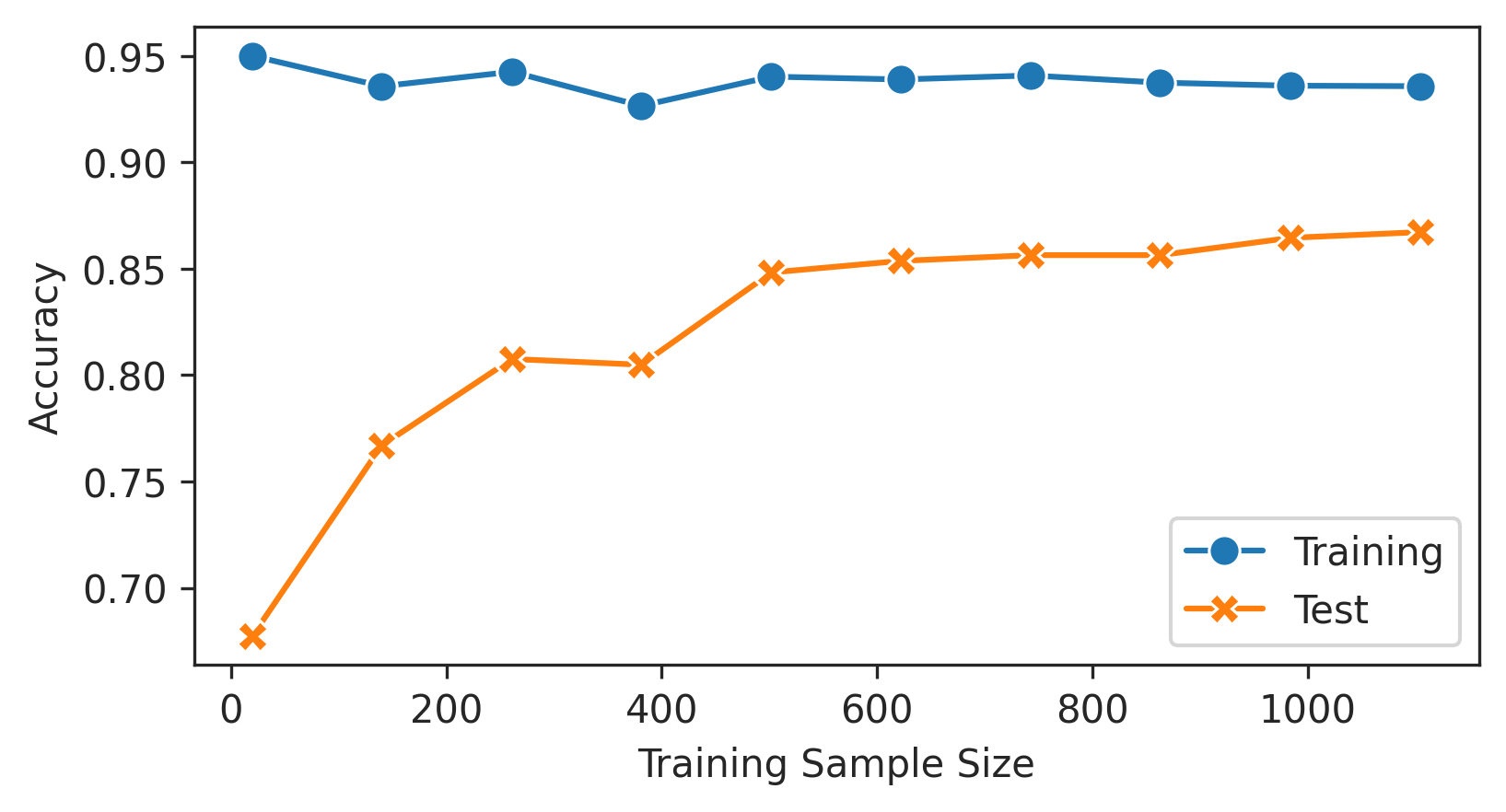}
        \end{subfigure}
        \caption{Data reduction analysis. (Top) A PCA cumulative variance plot of the TCGA-LGG data, showing that 121 dimensions are required to capture a 95\% variance threshold. (Bottom) A learning curve for the PhysioNet2017-NA data, generated with a classical proxy model (RBF-SVM), used to identify an efficient sample size where test accuracy begins to plateau (around 500 samples). \label{fig:analysis}}
    \end{figure}

    After these initial processes, an analysis of the data is crucial in order to properly reduce sample size and dimension (\Cref{fig:analysis}). First, to determine the feature dimension number, we conducted an explained variance analysis using Principal Component Analysis (PCA). The target dimension was selected to simultaneously satisfy the qubit-count constraints of our simulated device and preserve maximal data variance. This was achieved by choosing the number of components corresponding to the \enquote{elbow} of the cumulative variance plot, or alternatively, the number required to capture a predefined variance threshold (e.g., 90\%). Second, to determine an efficient training sample size, we generated a learning curve using a computationally inexpensive classical proxy model (the RBF kernel SVM). By plotting model performance against increasing sample sizes, we selected a sample size at which performance gains began to plateau within our computational budget. A similar rationale was applied to size the test set, ensuring it was large enough for statistically stable performance evaluation without incurring excessive simulation time during inference. We directly slice the datasets from the beginning to the target sample number for those that are already randomly split, while for the others, we performed stratified sampling using the \texttt{StratifiedShuffleSplit}. Then, we reduced the number of features using PCA.

    Finally, the features were scaled. We identified that the performance of the quantum kernel is highly sensitive to the sign of the input features. \texttt{StandardScaler} produces positive and negative feature values, which, after normalization, are encoded as positive and negative amplitudes. This corresponds to introducing relative phase information into the quantum state. In contrast, \texttt{MinMaxScaler} ensures all features are non-negative, thereby restricting all embedded data points to the positive orthant of the Hilbert space. Our preliminary tests showed that while classical kernels were negligibly affected, the quantum kernel's performance was significantly improved with \texttt{MinMaxScaler}, as the fidelity-based kernel is biased towards data representations in the positive orthant. Therefore, \texttt{MinMaxScaler} was chosen as the default scaler for subsequent comparative experiments, and \texttt{StandardScaler} for a few instances where it was more effective. The preprocessing procedures for each dataset are documented in \Cref{tab:preprocessing} in the Appendix.

    \subsubsection{Quantum Kernel Implementation}
    The quantum circuits introduced in \Cref{kernel_design} were simulated using the PennyLane library \cite{bergholmPennyLaneAutomaticDifferentiation2022}. The encoding of \texttt{QAmp} was handled by the built-in \texttt{AmplitudeEmbedding} function and \texttt{QRBF} was prepared by \texttt{StatePrep}. We employed PennyLane's \texttt{default.qubit} device with the JAX interface \cite{bradburyJAXComposableTransformations2018}. This setup was essential for efficiently executing our experiments, reducing the runtime of the extensive matrix calculations and gradient-based training required in our study. By using a statevector simulator, our calculations are based on the exact kernel values as defined in the infinite-shot limit.

    The quantum kernel matrix $\mathbf{K}$ quantifies the similarity between data points. The standard method for the \texttt{QAmp} kernel computes each element $K_{ij}$ from a single application of the kernel function $k$ to the complete data vectors $\mathbf{x}_i$ and $\mathbf{x}_j$. This provides a global measure of similarity between the two vectors. It is defined as
    \begin{equation} K_{ij} = k(\mathbf{x}_i, \mathbf{x}_j). \end{equation}

    The method for the \texttt{QRBF} kernel, which processes the feature elements, employs a component-wise decomposition. Here, the kernel function $k$ is first applied to each pair of corresponding features $(x_i^m, x_j^m)$ from the vectors $\mathbf{x}_i$ and $\mathbf{x}_j$. The final kernel element $K_{ij}$ is the arithmetic mean of these $d$ individual feature-wise kernel evaluations. This approach defines the total similarity as the average of similarities across all feature dimensions,
    \begin{equation} K_{ij} = \frac{1}{d} \sum_{m=1}^d k(x_i^m, x_j^m). \end{equation}

    This additive composition was chosen over a tensor product due to its superior performance observed in our experiments. This method effectively treats each feature independently for the overall similarity, in contrast to the entangled, holistic evaluation of the tensor kernel.

    The trainable parameters of the ansatz were initialized with a uniform distribution over $[0, 2\pi]$ and optimized by maximizing the kernel--target alignment $A(K,T)=\langle K,T\rangle_F/(\|K\|_F\|T\|_F)$ \cite{cristianiniKernelTargetAlignment2001} with the class-balanced target $T=\tilde{y}\tilde{y}^\top$, $\tilde{y}_i=\pm 1/n_{\pm}$ (first-vs-rest binarization for the three-class SEED-P12S1 task) \cite{hubregtsenTrainingQuantumEmbedding2022}, evaluated without centering. For optimization, each training set was further split into a 75\% sub-training set and a 25\% validation set. We employed the Adam optimizer with a learning rate of 0.05. The parameters were updated using mini-batches of size four for a total of 500 optimization steps. To select the optimal parameters, the KTA on the validation set was evaluated every 50 steps, and the parameter set that achieved the highest alignment over the entire training process was retained for the final kernel computation. The chosen hyperparameters (e.g., learning rate, batch size) are consistent with common practices and were refined during preliminary experiments.

    \subsubsection{Classification and Evaluation}
    The classification model was a support vector classifier (SVC) from scikit-learn. To evaluate general-purpose capability, two standard classical kernels, a linear kernel and an RBF kernel, were chosen as baselines for comparison with the quantum kernels within the same SVC framework. For all classical kernels, relevant hyperparameters (the SVC's regularization parameter \texttt{C} and the RBF kernel's \texttt{gamma}) were optimized using a grid search over a logarithmic scale.

    However, the hyperparameters for quantum kernels cannot be tuned by an exhaustive search due to the computational cost. We therefore designed a simple two-stage randomized search algorithm. The initial stage conducts a random search, sampling hyperparameters from curated sets of discrete values across a logarithmic scale. For the \texttt{QAmp} kernel, this includes \texttt{s} and the SVC's \texttt{C}. For the \texttt{QRBF}, the search also includes the gamma-like parameter \texttt{c}. After half of the total number of iterations, the algorithm transitions to a focused search. This stage samples new hyperparameter candidates from a continuous log-uniform distribution centered around the best-performing parameter set found in the broad search phase for \texttt{s} and \texttt{C}, while the \texttt{c} parameter continues to be sampled from its original discrete set throughout all iterations. For each sampled hyperparameter combination, the kernel's internal parameters were first optimized, and then an SVC was trained and evaluated. The final output is the hyperparameter set that achieved the highest validation score across all iterations. This method is efficient for finding the optimized hyperparameter combination in approximately 14 iterations for the \texttt{QAmp}, and 20 iterations for the \texttt{QRBF}. The final optimized hyperparameters are documented in \Cref{tab:hyperparameters} in the Appendix. All reported results use five random seeds $\{42, 0, 1, 2, 3\}$, with hyperparameters optimized once on seed 42 and held fixed thereafter. The performance of all kernel methods was evaluated using the classification accuracy on the unseen test sets and the macro-averaged F1-score, using the hyperparameters originally optimized for accuracy.

    \section{Results\label{sec:results}}
    This section presents our primary performance comparison, including an analysis of performance across different datasets, a validation of the scaling parameter technique, and an exploration of performance enhancement with respect to qubit resources.

    \begin{figure*}[htbp]
        \includegraphics[width=\linewidth]{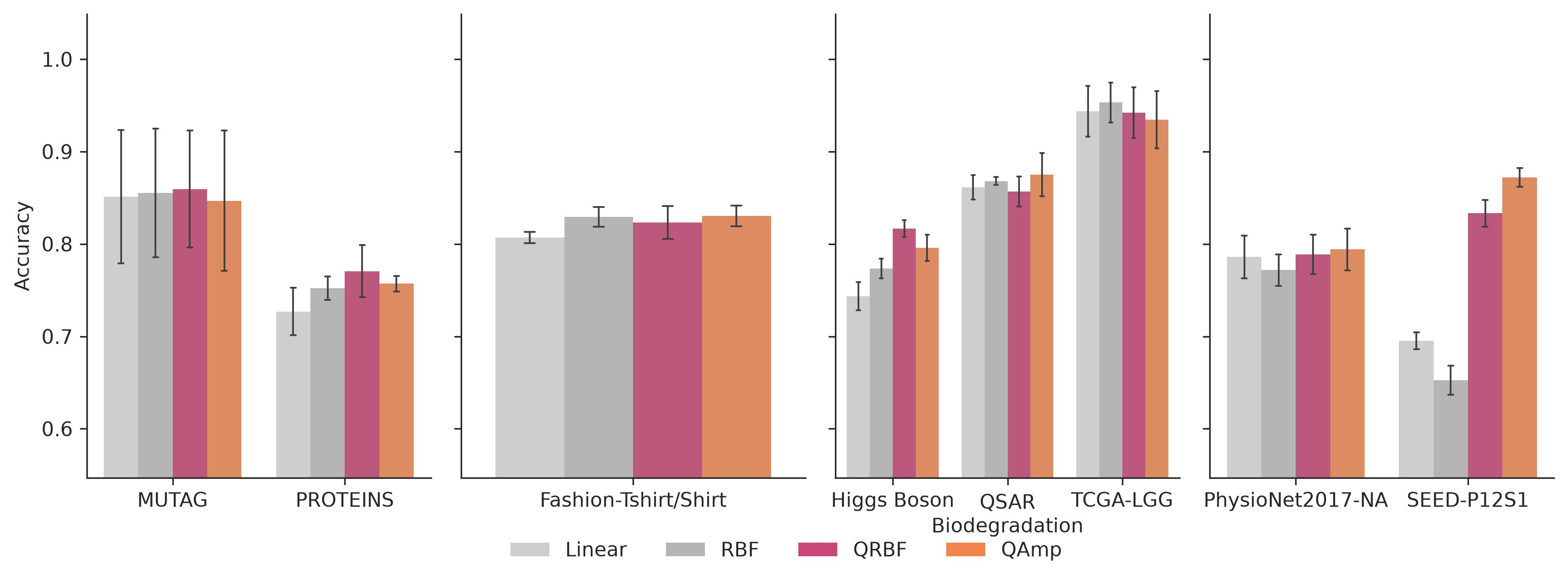}
        \caption{Comparison of classification accuracies of classical (Linear, RBF) and quantum (\texttt{QRBF}, \texttt{QAmp}) kernels, represented by light gray, dark gray, purple, and orange bars, respectively. Error bars show the sample standard deviation. Results are presented for eight benchmark datasets, categorized into graph, image, tabular, and time series data types, from left to right. \label{fig:results}}
    \end{figure*}

    \begin{table*}[htbp]
      \begin{tblr}{
        colspec = {l l l *{4}{c}},
        row{1-2} = {c, font=\bfseries},
        cell{3}{1} = {r=6}{m}, 
        cell{9}{1} = {r=2}{m}, 
        cell{11}{1} = {r=4}{m}, 
        cell{15}{1} = {r=4}{m}, 
        cell{3,5,7,9,11,13,15,17}{2} = {r=2}{m}, 
        cell{3-Z}{3} = {font=\itshape},
        hline{5,7,13,17} = {2-7}{0.5pt, solid, gray!30},
        }
        \toprule
        Type        & Dataset             & Metric & \SetCell[c=4]{c} {Performance (mean \(\pm\) std)} & & & \\
        \cmidrule[lr]{4-7}
                    &                     &        & {Linear}    & {RBF}       & {QRBF}      & {QAmp} \\
        \midrule
        Tabular     & Higgs Boson         & Acc   & $0.7434 \pm 0.0151$ & $0.7734 \pm 0.0108$ & $\mathbf{0.8166 \pm 0.0091}$ & $0.7958 \pm 0.0142$ \\
                    &                     & MF     & $0.7002 \pm 0.0171$ & $0.7402 \pm 0.0141$ & $\mathbf{0.7933 \pm 0.0116}$ & $0.7702 \pm 0.0200$ \\
                    & QSAR Biodegradation & Acc   & $0.8614 \pm 0.0133$ & $0.8682 \pm 0.0041$ & $0.8568 \pm 0.0164$ & $\mathbf{0.8750 \pm 0.0235}$ \\
                    &                     & MF     & $0.8433 \pm 0.0149$ & $0.8517 \pm 0.0078$ & $0.8404 \pm 0.0210$ & $\mathbf{0.8579 \pm 0.0250}$ \\
                    & TCGA-LGG            & Acc   & $0.9437 \pm 0.0273$ & $\mathbf{0.9531 \pm 0.0214}$ & $0.9422 \pm 0.0274$ & $0.9344 \pm 0.0311$ \\
                    &                     & MF     & $0.9100 \pm 0.0390$ & $\mathbf{0.9258 \pm 0.0314}$ & $0.9087 \pm 0.0377$ & $0.9030 \pm 0.0391$ \\
        \midrule
        Image       & Fashion-Tshirt/Shirt & Acc   & $0.8072 \pm 0.0061$ & $0.8292 \pm 0.0106$ & $0.8232 \pm 0.0176$ & $\mathbf{0.8304 \pm 0.0112}$ \\
                    &                     & MF     & $0.8071 \pm 0.0061$ & $0.8289 \pm 0.0108$ & $0.8228 \pm 0.0178$ & $\mathbf{0.8302 \pm 0.0113}$ \\
        \midrule
        Time Series & PhysioNet2017-NA    & Acc   & $0.7859 \pm 0.0231$ & $0.7718 \pm 0.0171$ & $0.7886 \pm 0.0214$ & $\mathbf{0.7940 \pm 0.0224}$ \\
                    &                     & MF     & $0.7856 \pm 0.0227$ & $0.7690 \pm 0.0181$ & $0.7884 \pm 0.0212$ & $\mathbf{0.7937 \pm 0.0225}$ \\
                    & SEED-P12S1          & Acc   & $0.6952 \pm 0.0092$ & $0.6524 \pm 0.0157$ & $0.8332 \pm 0.0146$ & $\mathbf{0.8720 \pm 0.0100}$ \\
                    &                     & MF     & $0.6819 \pm 0.0104$ & $0.6288 \pm 0.0177$ & $0.8290 \pm 0.0155$ & $\mathbf{0.8734 \pm 0.0104}$ \\
        \midrule
        Graph       & MUTAG               & Acc   & $0.8511 \pm 0.0722$ & $0.8553 \pm 0.0696$ & $\mathbf{0.8596 \pm 0.0631}$ & $0.8468 \pm 0.0758$ \\
                    &                     & MF     & $0.8223 \pm 0.0756$ & $0.8283 \pm 0.0710$ & $\mathbf{0.8353 \pm 0.0644}$ & $0.8185 \pm 0.0824$ \\
                    & PROTEINS            & Acc   & $0.7269 \pm 0.0258$ & $0.7520 \pm 0.0128$ & $\mathbf{0.7706 \pm 0.0283}$ & $0.7570 \pm 0.0086$ \\
                    &                     & MF     & $0.7080 \pm 0.0230$ & $0.7448 \pm 0.0117$ & $\mathbf{0.7619 \pm 0.0263}$ & $0.7488 \pm 0.0060$ \\
        \bottomrule
        \end{tblr}
        \caption{Comparison of kernel performance across diverse datasets. Entries are mean \(\pm\) standard deviation. \enquote{Acc} denotes accuracy; \enquote{MF} denotes the macro-averaged F1-score. The highest mean for each metric within a dataset is bolded. \label{tab:results}}
    \end{table*}

    \Cref{fig:results} and \Cref{tab:results} present the comparative classification accuracies of the four SVM kernels---classical Linear and RBF, quantum-native \texttt{QAmp} and classically inspired \texttt{QRBF}---across the eight benchmark datasets. The quantum kernels achieve classification accuracies comparable to those of the linear and RBF kernels, with higher mean accuracies on several datasets. The \texttt{QAmp} kernel holds the highest mean accuracy on Fashion-Tshirt/Shirt, QSAR Biodegradation, PhysioNet2017-NA, and most notably the SEED-P12S1 time series dataset, where it outperforms the best classical kernel by a large margin in all five seeds. The \texttt{QRBF} kernel achieves the highest mean on the PROTEINS and Higgs Boson datasets. Classical RBF has the highest mean on TCGA-LGG. MUTAG, Fashion-Tshirt/Shirt, and QSAR are near-ties across kernels. Paired accuracy differences and seed-wise comparison counts for all kernel contrasts are reported in \Cref{tab:paired,tab:paired-cont} in the Appendix.

\begin{figure}[htbp]
        \includegraphics[width=\linewidth]{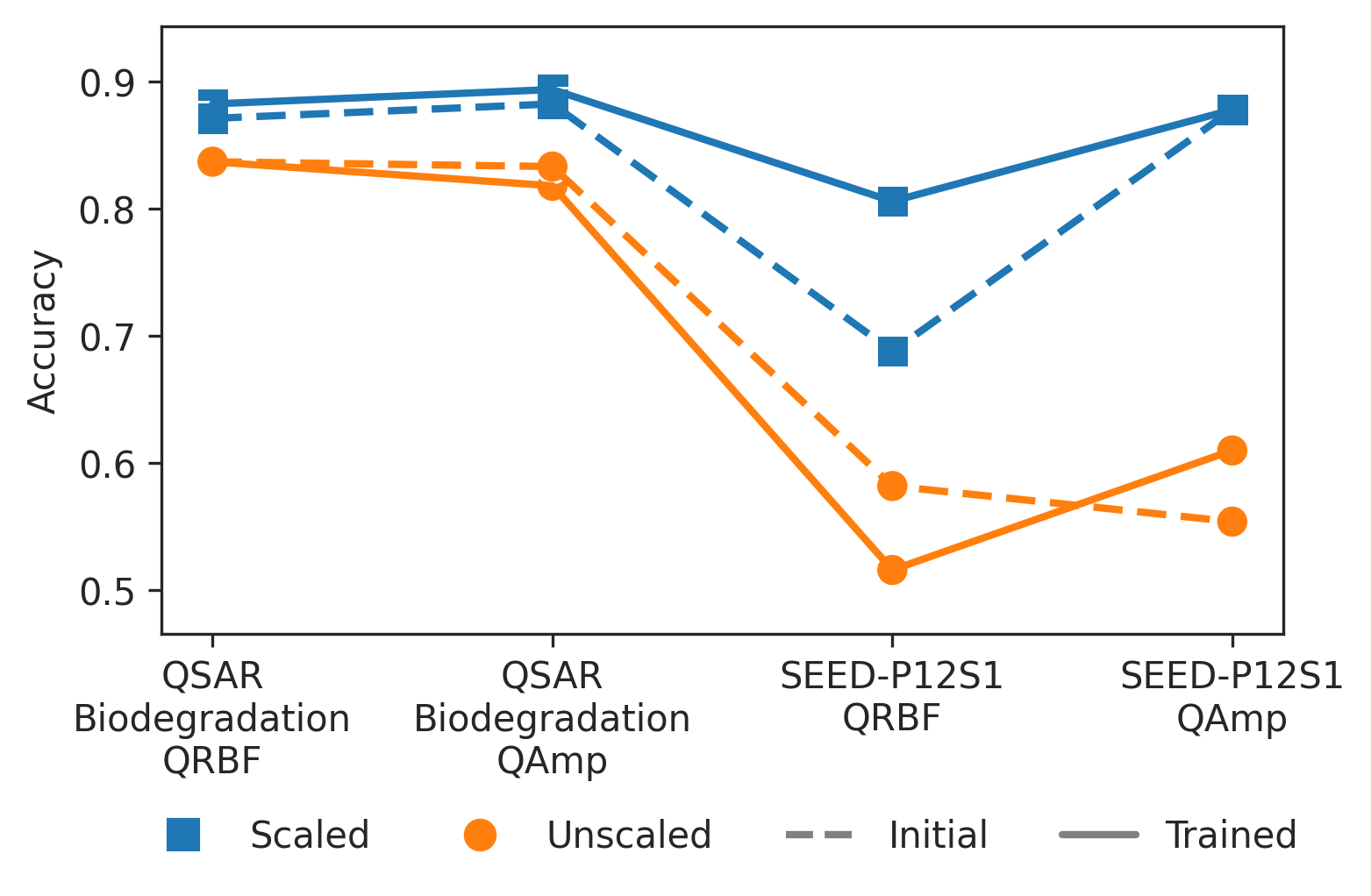}
        \caption{Comparison of initial (dashed) and trained (solid) accuracies for the ansätze with the scaling parameter $s$ (blue) and without it (orange). Results are shown for the \texttt{QRBF} and \texttt{QAmp} kernels on the QSAR Biodegradation and SEED-P12S1 datasets. \label{fig:results_s}}
    \end{figure}

    To evaluate the impact of the parameter scaling technique, we conducted a comparison between a standard, unscaled quantum ansatz and our proposed scaled ansatz over two representative datasets, the QSAR Biodegradation and SEED-P12S1, in \Cref{fig:results_s} (seed-42 configuration). First, we analyzed the performance of the standard, unscaled ansatz. \Cref{fig:results_s} shows that after training, its accuracy decreased in many cases. In contrast, the scaled ansatz demonstrated superior behavior: it not only achieved higher initial accuracy, but its performance was also enhanced or maintained for high-performance cases through training, resulting in a significantly higher final accuracy on both datasets.

\begin{figure}[htbp]
        \includegraphics[width=\linewidth]{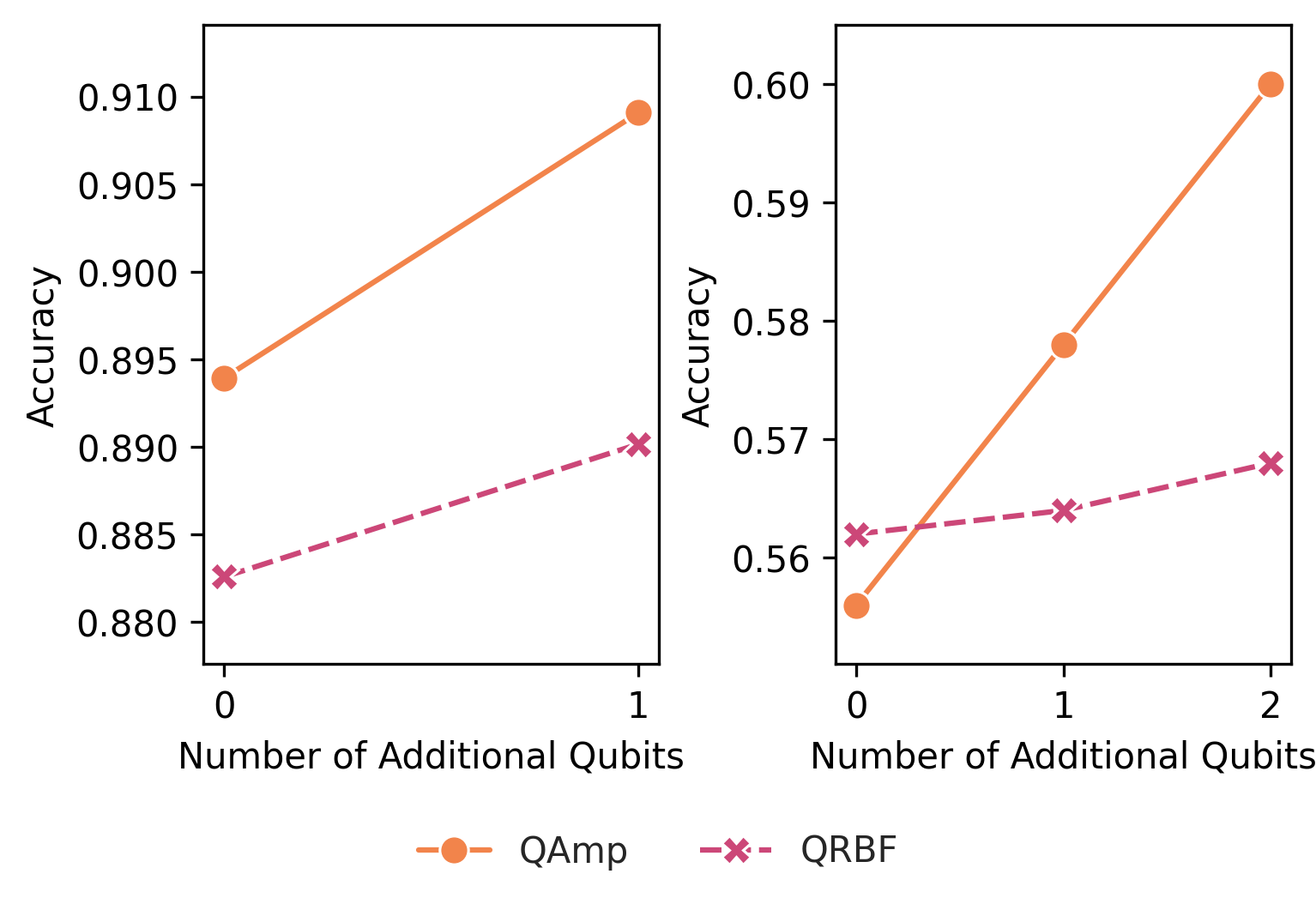}
        \caption{Effect of increasing qubit resource on the kernel accuracy. Results are shown for the QSAR Biodegradation dataset (left) and the SEED-P14S3 dataset (right). The baseline at zero additional qubits represents the original implementation. \label{fig:results_add}}
    \end{figure}

    As a final exploration, we assessed the potential for performance improvement by increasing the qubit count beyond the minimal requirements. For the \texttt{QAmp} kernel, this was achieved by re-uploading the data features across additional qubits. For the Quantum \texttt{QRBF} kernel, the modification involved both adding qubits and increasing the entanglement density by creating a local three-qubit interaction. Specifically, each qubit $i$ acts as a control for CNOTs on its subsequent two neighbors ($i+1$ and $i+2$) in a circular chain. The results for two representative datasets are presented in \Cref{fig:results_add} (seed-42 configuration).

    For the QSAR Biodegradation dataset, we evaluated the effect of adding one additional qubit. This modification led to a performance increase for both kernel types. A similar trend was observed in the SEED-P14S3 dataset (the third session of the participant 14), where we tested the addition of one and two qubits. The \texttt{QRBF} kernel showed a modest but steady increase in accuracy. The \texttt{QAmp} kernel, however, demonstrated substantial performance gains with each addition.

    \section{Discussion and Conclusion}
    We find that quantum kernels with parameter scaling achieve classification accuracies comparable to those of the linear and RBF kernels, with higher mean accuracies on several datasets. This finding provides a foundation for integrating quantum methods into real-world machine learning workflows in the future. These results suggest that the quantum models are versatile and can function as general-purpose kernels, similar to the classical RBF, positioning them as viable, high-dimensional candidates for future quantum-enhanced machine learning pipelines.

    Multiple kernel learning and metric learning automate kernel selection beyond discrete hyperparameter search \cite{gonenMultipleKernelLearning2011,jainMetricKernelLearning2012}; kernel--target alignment is a standard classical objective in that literature \cite{cristianiniKernelTargetAlignment2001}. Competitive performance with classical kernels, with gains on a subset of datasets, is consistent with the inductive-bias analysis of quantum kernels, under which advantages on classical data require an inductive bias matched to the data-generating process \cite{kublerInductiveBiasQuantum2021}.

    Our results in \Cref{fig:results_s} validate the parameter scaling technique, which provides a practical strategy for accelerating convergence by addressing both the weak initial expressivity and unstable training that affect many variational quantum algorithms. This, alongside other techniques and experience throughout the ML process, can serve as a methodology to advance future QML for real-world tasks. The observation of performance enhancement with increasing qubit count in our experiments suggests that increasing the number of qubits used for data encoding and entanglement may enhance the classification accuracy of quantum kernels by utilizing entanglement generated by the PQC. Increasing the qubit count could improve classification accuracy, although exponential concentration and vanishing similarities may limit this improvement \cite{thanasilpExponentialConcentrationQuantum2024, suzukiQuantumFisherKernel2024}.

    As argued in \cite{peral-garciaSystematicLiteratureReview2024}, the full potential of the QSVM algorithm, which is tied to its ability to take advantage of an exponentially large $2^N$-dimensional space, is expected to emerge after the noisy intermediate-scale quantum (NISQ) era. Given the intrinsic limitations of SVM methods, which depend on pre-processed, structured feature vectors and given their prohibitive computational complexity when applied to large-scale datasets, along with their efficiency for small to medium-sized datasets, this makes QSVM a promising tool specifically for such datasets, particularly when they are high-dimensional. As we approach the fault-tolerant quantum computing (FTQC) era, we envision the quantum kernel playing a role similar to that of linear or RBF kernels in SVM in the classical machine learning world: an effective classification tool for high-dimensional, moderate-scale datasets for which large-scale, end-to-end architectures like neural networks are not required or may even underperform. We anticipate that future research will uncover further benefits unique to quantum kernels. While our circuit design uses effective and general settings to reduce resource costs, as devices with higher resources become available, researchers could develop more complex and potentially powerful designs to further enhance performance, building on the baseline from this work. We hypothesize that the primary advantage of quantum kernels is expected for problems with extreme feature dimensions ($>10^6$, and particularly for $>10^9$) and a moderate number of samples ($\approx 10^4-10^5$). This would leverage the quantum computer's strength in high-dimensional spaces, while ensuring the classical SVM optimization, which scales with the sample size, remains manageable \cite{liuRigorousRobustQuantum2021}.

    Our study purposefully focused on four fundamental data structures---vectors, grids, sequences, and graphs---to provide a baseline for quantum kernels. We consciously excluded more complex or combined data types to avoid confounding variables and maintain clarity in our results. For instance, modern text (NLP) analysis relies heavily on large classical embedding models, and including them would test the embedding more than the kernel. Similarly, audio and video data can be seen as complex examples of time series and spatiotemporal sequences, respectively. Understanding the kernel's performance on the fundamental building blocks (images and time series) is a crucial first step before studying their combination. These more complex data types---including multimodal fusion---represent important and logical next steps that build directly upon the foundational results of this work.

    While the datasets such as MUTAG and PROTEINS were chosen to represent complex data structures, they are of a modest scale compared to the massive datasets common in industrial deep learning, which is due to the inherent limitations of vector kernel methods. Indeed, this small-to-medium scale tabular domain, long dominated by tree-based models, is now seeing the rise of powerful, end-to-end learned foundation models like TabPFN that achieve state-of-the-art performance with zero-shot inference \cite{hollmannAccuratePredictionsSmall2025}. This rapid classical progress underscores the need for more quantum-native approaches beyond simple vectorized kernels, such as the structure-aware quantum feature maps which can embed the structure of graphs, sequences, or images directly into the circuit's architecture \cite{wazniLargeScaleStructureaware2024, tangQuantumSelfattentionVisual2025}. While practical implementation is restricted by today's hardware, these methods represent a much more quantum-native approach to machine learning and may be a key area where quantum computers could eventually show a real advantage.

    Furthermore, our results were obtained in the idealized infinite-shot limit. A next step is to determine the performance on real quantum devices where the kernel matrix is statistically estimated from a finite number of circuit executions. In addition, future efforts could explore the exponential speed-up potential of quantum kernels over classical ones for real-world, classically challenging problems. This would demonstrate a computational advantage for real-world applications, building on theoretical work that has proven such speed-ups are possible for specifically constructed problems \cite{liuRigorousRobustQuantum2021}.

    In conclusion, we have designed and systematically validated a resource-efficient quantum kernel architecture that performs on par with classical methods on a diverse set of problems. By introducing a practical technique for accelerating convergence and maximizing qubit efficiency, this work provides a blueprint for the future application of quantum kernel methods. As hardware continues to improve and more sophisticated, structure-aware embeddings become feasible, the principles and designs presented here will serve as a basis for addressing increasingly complex challenges and realizing the full potential of quantum machine learning.

    \section*{Acknowledgments}
    This work was supported by funding from the NSF QLCI for Hybrid Quantum Architectures and Networks (NSF award 2016136).

    \section*{Data Availability}
    The code and data that support the findings of this study are available from the corresponding author on reasonable request.

    \section*{Author Contributions}
    M.~O. and J.~Y. conceptualized the project. J.~Y. wrote and ran the simulations. All authors contributed to the research and wrote the paper.

    \clearpage
    \onecolumngrid

    \appendix*
    \section{Experimental Details}
    \begin{table}[h!]
        \caption{Summary of dataset preprocessing procedures. \label{tab:preprocessing}}
        \begin{tblr}{row{1} = {font=\bfseries}}
        \toprule
        Dataset & Preprocessing Steps & Dimension Reduction & Feature Scaling \\
        \midrule
        MUTAG & Feature engineering & \textemdash & \texttt{MinMaxScaler} \\
        PROTEINS & Feature engineering & \textemdash & \texttt{MinMaxScaler} \\
        Fashion-Tshirt/Shirt & \textemdash & {700 train / 500 test samples \\ 128 features} & \texttt{StandardScaler} \\
        Higgs Boson & Median impute & 2500 train / 625 test samples & \texttt{MinMaxScaler} \\
        QSAR Biodegradation & \textemdash & \textemdash & \texttt{StandardScaler} \\
        TCGA-LGG & \textemdash & 128 features & \texttt{MinMaxScaler} \\
        PhysioNet2017-NA & {Class balancing \\ Signal truncation \\ Feature engineering} & {500 train samples \\ 128 features} & \texttt{MinMaxScaler} \\
        SEED-P12S1 & Feature engineering & 700 train / 500 test samples & \texttt{MinMaxScaler} \\
        \bottomrule
        \end{tblr}
    \end{table}

    \begin{table}[h!]
	\caption{Optimal hyperparameters for each kernel across all datasets. Long floating-point values were rounded to three decimal places. \label{tab:hyperparameters}}
        \begin{tblr}{row{1} = {font=\bfseries}}
            \toprule
            Dataset & Kernel & Hyperparameters \\
            \midrule
            \SetCell[r=4]{l} MUTAG & Linear & $C$: 1 \\
            & RBF & $C$: 10, $\gamma$: \texttt{auto}  \\
            & QRBF & $c$: 0.707, $s$: 0.05, $C$: 100 \\
            & QAmp & $s$: 0.02, $C$: 100 \\
            \addlinespace
            \SetCell[r=4]{l} PROTEINS & Linear & $C$: 100 \\
            & RBF & $C$: 100, $\gamma$: 1 \\
            & QRBF & $c$: 0.707, $s$: 0.75, $C$: 10 \\
            & QAmp & $s$: 1, $C$: 10  \\
            \addlinespace
            \SetCell[r=4]{l} Fashion-Tshirt/Shirt & Linear & $C$: 1 \\
            & RBF & $C$: 1, $\gamma$: \texttt{scale}  \\
            & QAmp & $s$: 0.0075, $C$: 1 \\
            & QRBF & $c$: \texttt{scale}, $s$: 0.001, $C$: 100 \\
            \addlinespace
            \SetCell[r=4]{l} Higgs Boson & Linear & $C$: 10  \\
            & RBF & $C$: 100, $\gamma$: 0.1 \\
            & QRBF & $c$: 2.236, $s$: 0.2, $C$: 100 \\
            & QAmp & $s$: 0.01, $C$: 1000  \\
            \addlinespace
            \SetCell[r=4]{l} QSAR Biodegradation & Linear & $C$: 0.1 \\
            & RBF & $C$: 10, $\gamma$: \texttt{auto} \\
            & QAmp & $s$: 0.0075, $C$: 1 \\
            & QRBF & $c$: 2.236, $s$: 0.075, $C$: 10  \\
            \addlinespace
            \SetCell[r=4]{l} TCGA-LGG & Linear & $C$: 0.1 \\
            & RBF & $C$: 1, $\gamma$: 0.1 \\
            & QRBF & $c$: 2.236, $s$: 0.002, $C$: 1000 \\
            & QAmp & $s$: 0.05, $C$: 100  \\
            \addlinespace
            \SetCell[r=4]{l} PhysioNet2017-NA & Linear & $C$: 10  \\
            & RBF & $C$: 1, $\gamma$: \texttt{scale} \\
            & QRBF & $c$: 0.224, $s$: 0.0075, $C$: 1000  \\
            & QAmp & $s$: 0.0075, $C$: 100 \\
            \addlinespace
            \SetCell[r=4]{l} SEED-P12S1 & Linear & $C$: 0.1 \\
            & RBF & $C$: 0.1, $\gamma$: \texttt{scale}  \\
            & QRBF & $c$: 2.236, $s$: 0.005, $C$: 1000 \\
            & QAmp & $s$: 0.002, $C$: 10 \\
            \bottomrule
        \end{tblr}
    \end{table}

    \begin{table}[h!]
        \caption{Paired accuracy differences (tabular and image datasets). For each contrast $A-B$, both kernels share the same seed. Entries are the mean \(\pm\) standard deviation of the paired differences; positive values favor $A$. The final column lists the number of runs with $A{>}B$, $A{=}B$, and $A{<}B$. Macro-averaged F1-score differences follow the same qualitative pattern. \label{tab:paired}}
        \begin{tblr}{
            colspec = {l l c c},
            row{1} = {font=\bfseries},
            hline{8,14,20} = {1-4}{0.5pt, solid, gray!30},
            }
            \toprule
            Dataset & Contrast & Mean \(\pm\) std & $A{>}B$ / $A{=}B$ / $A{<}B$ \\
            \midrule
            Higgs Boson & \texttt{QAmp}$-$Linear & $+0.0525 \pm 0.0280$ & 5 / 0 / 0 \\
            & \texttt{QAmp}$-$RBF & $+0.0224 \pm 0.0161$ & 5 / 0 / 0 \\
            & \texttt{QRBF}$-$Linear & $+0.0733 \pm 0.0143$ & 5 / 0 / 0 \\
            & \texttt{QRBF}$-$RBF & $+0.0432 \pm 0.0060$ & 5 / 0 / 0 \\
            & \texttt{QAmp}$-$\texttt{QRBF} & $-0.0208 \pm 0.0185$ & 0 / 1 / 4 \\
            & RBF$-$Linear & $+0.0301 \pm 0.0164$ & 5 / 0 / 0 \\
            QSAR Biodegradation & \texttt{QAmp}$-$Linear & $+0.0136 \pm 0.0205$ & 3 / 1 / 1 \\
            & \texttt{QAmp}$-$RBF & $+0.0068 \pm 0.0196$ & 3 / 1 / 1 \\
            & \texttt{QRBF}$-$Linear & $-0.0045 \pm 0.0233$ & 2 / 0 / 3 \\
            & \texttt{QRBF}$-$RBF & $-0.0114 \pm 0.0131$ & 1 / 0 / 4 \\
            & \texttt{QAmp}$-$\texttt{QRBF} & $+0.0182 \pm 0.0174$ & 4 / 0 / 1 \\
            & RBF$-$Linear & $+0.0068 \pm 0.0130$ & 4 / 0 / 1 \\
            TCGA-LGG & \texttt{QAmp}$-$Linear & $-0.0094 \pm 0.0128$ & 1 / 1 / 3 \\
            & \texttt{QAmp}$-$RBF & $-0.0187 \pm 0.0188$ & 1 / 0 / 4 \\
            & \texttt{QRBF}$-$Linear & $-0.0016 \pm 0.0065$ & 1 / 2 / 2 \\
            & \texttt{QRBF}$-$RBF & $-0.0109 \pm 0.0105$ & 0 / 2 / 3 \\
            & \texttt{QAmp}$-$\texttt{QRBF} & $-0.0078 \pm 0.0096$ & 1 / 0 / 4 \\
            & RBF$-$Linear & $+0.0094 \pm 0.0086$ & 4 / 1 / 0 \\
            Fashion-Tshirt/Shirt & \texttt{QAmp}$-$Linear & $+0.0232 \pm 0.0135$ & 5 / 0 / 0 \\
            & \texttt{QAmp}$-$RBF & $+0.0012 \pm 0.0120$ & 3 / 0 / 2 \\
            & \texttt{QRBF}$-$Linear & $+0.0160 \pm 0.0212$ & 3 / 0 / 2 \\
            & \texttt{QRBF}$-$RBF & $-0.0060 \pm 0.0133$ & 2 / 0 / 3 \\
            & \texttt{QAmp}$-$\texttt{QRBF} & $+0.0020 \pm 0.0068$ & 2 / 1 / 2 \\
            & RBF$-$Linear & $+0.0220 \pm 0.0161$ & 4 / 0 / 1 \\
            \bottomrule
        \end{tblr}
    \end{table}

    \begin{table}[h!]
        \caption{Paired accuracy differences (time series and graph datasets). Conventions as in \Cref{tab:paired}. \label{tab:paired-cont}}
        \begin{tblr}{
            colspec = {l l c c},
            row{1} = {font=\bfseries},
            hline{8,14,20} = {1-4}{0.5pt, solid, gray!30},
            }
            \toprule
            Dataset & Contrast & Mean \(\pm\) std & $A{>}B$ / $A{=}B$ / $A{<}B$ \\
            \midrule
            PhysioNet2017-NA & \texttt{QAmp}$-$Linear & $+0.0081 \pm 0.0090$ & 3 / 2 / 0 \\
            & \texttt{QAmp}$-$RBF & $+0.0222 \pm 0.0169$ & 5 / 0 / 0 \\
            & \texttt{QRBF}$-$Linear & $+0.0027 \pm 0.0227$ & 3 / 0 / 2 \\
            & \texttt{QRBF}$-$RBF & $+0.0168 \pm 0.0075$ & 5 / 0 / 0 \\
            & \texttt{QAmp}$-$\texttt{QRBF} & $+0.0022 \pm 0.0222$ & 2 / 0 / 3 \\
            & RBF$-$Linear & $-0.0141 \pm 0.0213$ & 1 / 1 / 3 \\
            SEED-P12S1 & \texttt{QAmp}$-$Linear & $+0.1768 \pm 0.0046$ & 5 / 0 / 0 \\
            & \texttt{QAmp}$-$RBF & $+0.2196 \pm 0.0062$ & 5 / 0 / 0 \\
            & \texttt{QRBF}$-$Linear & $+0.1380 \pm 0.0084$ & 5 / 0 / 0 \\
            & \texttt{QRBF}$-$RBF & $+0.1808 \pm 0.0061$ & 5 / 0 / 0 \\
            & \texttt{QAmp}$-$\texttt{QRBF} & $+0.0388 \pm 0.0076$ & 5 / 0 / 0 \\
            & RBF$-$Linear & $-0.0428 \pm 0.0074$ & 0 / 0 / 5 \\
            MUTAG & \texttt{QAmp}$-$Linear & $-0.0043 \pm 0.0233$ & 1 / 3 / 1 \\
            & \texttt{QAmp}$-$RBF & $-0.0085 \pm 0.0323$ & 1 / 3 / 1 \\
            & \texttt{QRBF}$-$Linear & $+0.0085 \pm 0.0117$ & 2 / 3 / 0 \\
            & \texttt{QRBF}$-$RBF & $+0.0043 \pm 0.0095$ & 1 / 4 / 0 \\
            & \texttt{QAmp}$-$\texttt{QRBF} & $-0.0128 \pm 0.0285$ & 0 / 4 / 1 \\
            & RBF$-$Linear & $+0.0043 \pm 0.0095$ & 1 / 4 / 0 \\
            PROTEINS & \texttt{QAmp}$-$Linear & $+0.0301 \pm 0.0226$ & 5 / 0 / 0 \\
            & \texttt{QAmp}$-$RBF & $+0.0050 \pm 0.0153$ & 4 / 0 / 1 \\
            & \texttt{QRBF}$-$Linear & $+0.0437 \pm 0.0053$ & 5 / 0 / 0 \\
            & \texttt{QRBF}$-$RBF & $+0.0186 \pm 0.0180$ & 4 / 0 / 1 \\
            & \texttt{QAmp}$-$\texttt{QRBF} & $-0.0136 \pm 0.0265$ & 2 / 0 / 3 \\
            & RBF$-$Linear & $+0.0251 \pm 0.0170$ & 5 / 0 / 0 \\
            \bottomrule
        \end{tblr}
    \end{table}

    \FloatBarrier
    \clearpage
    \twocolumngrid
    \bibliography{references}
\end{document}